\documentclass[11pt]{article}
\usepackage{graphicx} % Required for inserting images
\usepackage[style=apa, backend=biber, natbib]{biblatex}
\usepackage[margin=1in]{geometry}
\usepackage{algorithm}
\usepackage{algpseudocode}
\usepackage{lscape}

\usepackage{color}

\title{Generative AI for Business Strategy: Using Foundation Models to Create Business Strategy Tools}
\author{Son The Nguyen and Theja Tulabandhula}

 \date{%
    University of Illinois at Chicago%
 }
\begin{document}

\maketitle

\begin{abstract}
Generative models (foundation models) such as LLMs (large language models) are having a large impact on multiple fields. In this work, we propose the use of such models for business decision making. In particular, we combine unstructured textual data sources (e.g., news data) with multiple foundation models (namely, GPT4, transformer-based Named Entity Recognition (NER) models and Entailment-based Zero-shot Classifiers (ZSC)) to derive IT (information technology) artifacts in the form of a (sequence of) signed business networks. We posit that such artifacts can inform business stakeholders about the state of the market and their own positioning as well as provide quantitative insights into improving their future outlook.
\end{abstract}
% The study emphasizes the importance of effectively managing conflict and collaboration relationships in modern businesses and demonstrates how news content analysis can provide insights into these relationships. Automatic extraction of entities and inference of relationships from news content allows businesses to gain a comprehensive understanding of the signed network of firms over time. However, previous studies have mainly focused on static and one-dimensional networks, which fail to adequately reflect the dynamic nature of the business environment. This paper presents a methodology for building a signed conflict-collaboration network of firms based on news content and context,  for conflict-collaboration relationship inference.

\section{Introduction}

In the fast-paced and dynamic world of modern business, companies face constant challenges from fierce competition and the need to innovate to remain relevant and competitive continuously. To succeed in this environment, companies often engage in a delicate interplay of collaboration and competition across various domains. Collaboration can be instrumental in sharing resources and expertise, leveraging strengths, and fostering mutually beneficial outcomes. However, companies also frequently encroach on each other's territory, seeking to establish dominance in their respective markets. This can lead to heightened competition and conflict, as companies jostle for an advantageous position in the market. This interplay between collaboration and competition is not static but rather evolving, as companies seek to build and protect their dynasties. The actions of one company can trigger a response from another, leading to a constantly shifting landscape of alliances and rivalries. In this work, we seek ways to employ foundation models to help business effectively navigate such market dynamics

% \subsection{Signed Business Networks}

Companies should carefully analyze and categorize their relationships with other firms on both a local and global level, recognizing the interconnectedness of different relationships across the industry. Thus, a network abstraction that denotes (time varying) relationships between entities in a market can be useful for such strategic decision making. We call such networks, \emph{signed business networks}, where the nodes correspond to entities (firms/products) and weighted edges represent their relationships with ohter entities.

% \subsection{Using Generative AI for Business Decision Making}

Multiple stakeholders such as the firm leadership, financial analysts, investors and policymakers can make use of a graphical abstraction. Using unstructured data, such as news content, we show that one can build such an abstraction with near zero necessity of technical expertise.

\section{Text Data Sources and Foundation Models}

One powerful tool for gaining insight into these relationships is by analyzing news content, which contains a wealth of information about the connections between companies. By automatically extracting entities and inferring relationships from headlines, descriptions, summaries, and other content, it's possible to obtain a comprehensive and detailed view of the signed network of firms over time.

When building a business network using natural language processing, the primary task is to identify the relationships between different organizations. Past studies have mainly relied on news as the primary data source and employed conventional techniques such as lexical patterns (\cite{Lau2011SemisupervisedSI}) or dependency parsing (\cite{Braun2018AutomaticRE}) to perform business relation extraction. However, advanced machine learning techniques like Markov Logic Network (\cite{7889512}), Bi-directional Gated Recurrent Units (\cite{8679499}), Conditional Random Field (\cite{10.1007/978-3-030-41505-1_24}), Bidirectional Encoder Representations from Transformers (BERT) (\cite{de-los-reyes-etal-2021-related}), and Multitask Models with Shared Sentence Encoders like BizBERT (\cite{khaldi:hal-03730345}) have been implemented recently. The majority of networks discussed in previous studies are static and fail to account for the various aspects of relationships, resulting in an inaccurate reflection of the dynamic nature of business interactions. 

This work proposes to build a sequence of signed conflict-collaboration network of firms based on news content \emph{using foundation models}. In particular, this will be accomplished by extracting organizations using a transformer-based Named Entity Recognition (NER) model and inferring their conflict-collaboration relationships using an Entailment-based Zero-shot Classifier (ZSC) to identify company relationships without any training automatically. Our approach will be augmented with GPT4 for textual explanations of the signs/weights of the learned network.

\subsection{Foundation Models for Text} \label{sec:preliminaries}
\subsubsection{Transformer-based Named Entity Recognizer}
Named entity recognition (NER) is the natural language processing task of detecting real-world entities in unstructured text and categorizing them to categories such as names of people, organizations, locations, times, quantities, monetary values, percentages, and more. An NER machine learning (ML) model, for example, may recognize the phrase ``University of Illinois at Chicago" in a text and classify it as an ``Organization.''

There are three ways to named entity recognition: dictionary-based, rule-based, and machine learning-based. A dictionary-based or rule-based methods maintain as many named entities and their vocabularies or the patterns of them appearing in real sentences as feasible in a database. These methods appear to be straightforward, yet they has drawbacks. Due to the ever-evolving nature of human language, it's practically difficult to develop an exhaustive list of entities or rules. New concepts or rules are constantly presented in a dynamic research sector like machine learning or biomedical field. Moreover, merged terms, term diversity or unique synonyms, popular English words overlapping with entity names, and confusing abbreviations complicate the compilation of a complete lexicon. In contrast, the machine learning-based technique known as deep learning may attach named entities to words without dictionaries or rules predefined and achieve state-of-the-art NER result. A transformer-based pretrained language model are fine-tuned to classify entity types on text corpus where the interested named entities are masked.

\subsubsection{Zero-shot Classifier}
Zero-shot Classifier proposed by (Yin et al. 2019) is a deep learning classifier that effectively predicts the class label without any prior training data pertaining to that label. Compared to traditional supervised learning approaches, which rely on a large number of examples for each class, the critical idea of ZSC
is based on the semantic transfer of information from observed labels to newly seen labels. ZSC uses a Natural Language Inference (NLI) model, which is a pre-trained sequence-pair transformer/classifier that uses both a premise and a hypothesis input to predict whether the hypothesis is true (entailment), false (contradiction), or undetermined (neutral) given the premise.

\subsubsection{GPT4 and other Instruction-tuned LLMs}

Generative Pre-trained Transformer 4 (GPT-4) is a multimodal large language model (LLM) created by the firm OpenAI.  As a transformer based model, GPT-4 was pretrained to predict the next token/word, and was then fine-tuned with reinforcement learning from human and AI feedback for human alignment and policy compliance. This is called instruction-tuning. Other models such as Claude, Bard, LLaMA belong to this category.

\section{Building Business Networks using Foundation Models} 

\begin{figure}[h]
\centering
\includegraphics[scale=0.55]{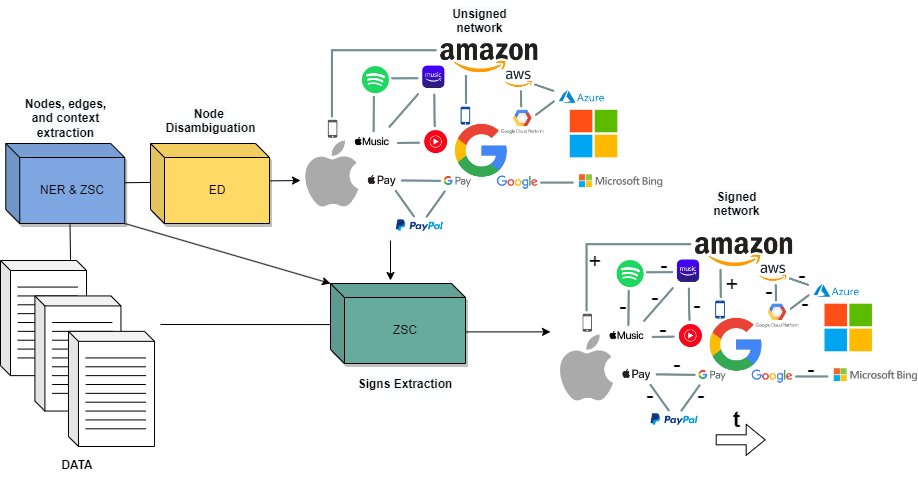}
\caption{Computing signed business networks using multiple foundation models.}
\label{fig:method-1}
\end{figure}

\begin{figure}[h]
\centering
\includegraphics[width=.8\textwidth]{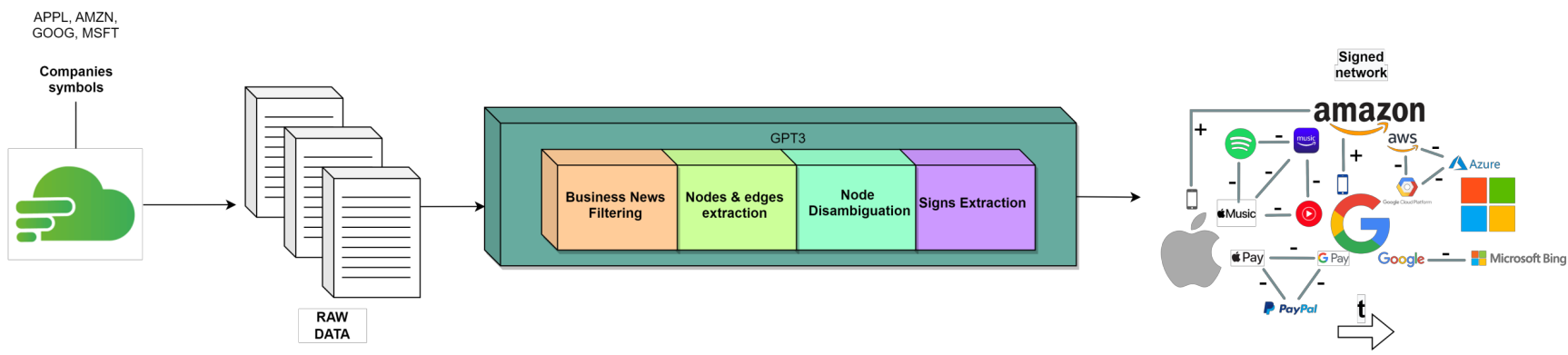}
\caption{We also build a parallel pipeline using Instruction-tuned LLMs. The goal here is to augment the signed network with expanations.}
\label{fig:method-2}
\end{figure}

To build our business network, we obtain company news using the News API. The News API provides 20 years of company news, including headlines, summaries, published dates, images, sources, and URLs. We build a database by feeding a set of interested firms Google, Amazon, Facebook, Apple, and Microsoft with their listed symbols through the API. Later on, we can scale it up to analyze all companies in a specific industry, the S{\&}P500, or all companies available in the News API. By using the News API and feeding in the interested companies, the API acts as a black box NER model that returns business news about the selected companies. We excluded stock news since financial or stock headlines alone do not provide sufficient context regarding the relationships between firms, which can result in inaccurate results. Therefore, we utilized a ZSC to filter out stock news.

Once we have obtained the data, our initial task is to identify the entities (nodes) and relationships (edges) by analyzing news headlines that mention two or more companies. To accomplish this, we employ the XLM-RoBERTa-large model, which is fine-tuned with the conll2003 dataset in English (\cite{conneau2019unsupervised}), to detect and extract organizations. Since companies may have various names, we utilize entity linking or entity disambiguation techniques to merge these different names into a single entity.

Because the relationships between organizations can be complex and multifaceted, we employ an additional ZSC to extract further details such as specific product lines, services, aspects, or fields mentioned in the news. The prediction, referred to as $<$Context$>$, is then combined with the NER model to create sub-entities associated with the primary company. These sub-entities establish direct connections with other sub-entities linked to different companies. Additionally, this information is used to generate prompts for our downstream relationship extraction models. The outcome of this step is an unsigned network of firms.

Once we have obtained the unsigned network, we proceed to determine the types of relationships using the ZSC Bart-Large-Mnli (\cite{yin-etal-2019-benchmarking}). The classifier assesses whether a hypothesis supports or contradicts a premise. In this instance, the premise is a news headline, and the hypothesis is as follows: ``the relationship between company A and B is $<$CLASS$>$," where the classes include ``positive", ``negative", ``neutral," and ``unknown."

% Currently, we assume that all news predictions carry equal weight, although this is an oversimplification. To determine the sign of a relationship, we utilize a majority vote based on the last ten occurrences. The generation of the signed network is outlined in Figure 1 to provide an overview of our methodology.

Complementing this pipeline, as shown in Figure~\ref{fig:method-2}, we use an instruction-tined LLM to provide explanations of the signs inferred from each news headline/article. These are then summarized after aggregating multiple news headlines/articles for any pair of entities.

\section{Preliminary Results} 

To demonstrate the feasibility of our approach, we show example analytical results from several gathered headlines using both the transformer based ZSC-NER-ZSC pipeline and the GPT4 pipeline. In most cases, we accurately identified the firms and the signs of the relationships (Tables 1 and 2). Even with more complex sentences, such as the one in row 3 of Table 1, our ZSC correctly inferred the positive relationship between Apple and Google. The same is true for GPT4. In comparison, the relation extraction model presented in (\cite{7889512}) was unsuccessful in classifying the relationship due to the term compete having a heavyweight in the n-gram feature.

\begin{figure}[h]
\centering
\includegraphics[width=.7\textwidth]{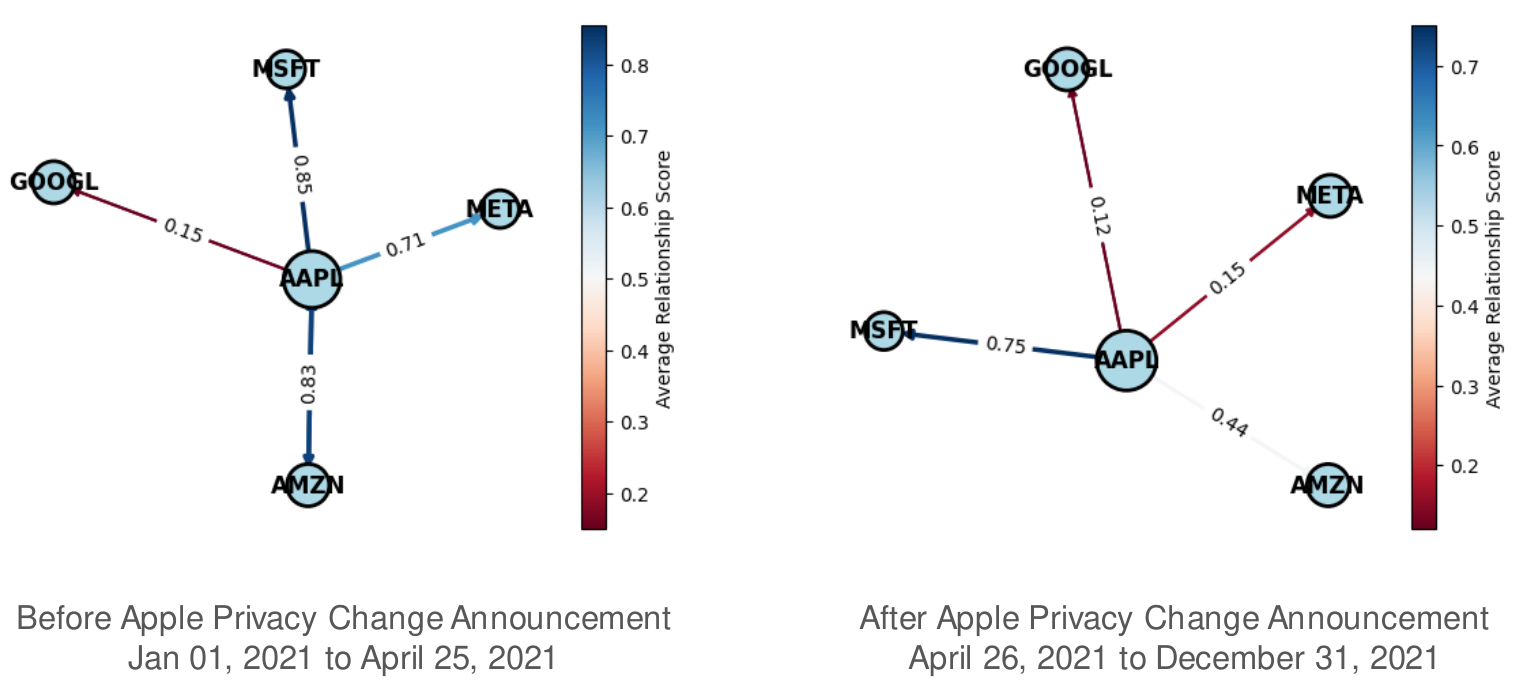}
\caption{Illustrative example: network changes before and after a business strategy change by Apple Inc., identified in a completely data-driven way using foundation models.}
\label{fig:apple-example}
\end{figure}

We also provide an illustrative example of how the entity relationships, learned in a fully data-driven way, reflect the business landscape via the firm Apple Inc. Apple Inc. changed its privacy policy for its App Store, negatively affecting multiple advertising focused companies. This is automatically reflected in the (sub-)network snapshots shown in Figure~\ref{fig:apple-example}. The example show that firms can obtain useful quantitative information regarding the markets they participate in, potentially improving their their business decision making approaches.

\begin{table}[h]
\centering
\begin{tabular}{|l|l|l|ll}
\cline{1-3}
\textbf{News headlines} &
  \textbf{NER result} &
  \textbf{\begin{tabular}[c]{@{}l@{}}ZSC result \\ (3 classes: positive, negative, \\ neutral)\end{tabular}} &
   &
   \\ \cline{1-3}
\begin{tabular}[c]{@{}l@{}}Facebook Paid GOP Firm \\ to Malign Tiktok\end{tabular} &
  \begin{tabular}[c]{@{}l@{}}Facebook, score: 0.99; \\ GOP Firm, score: 0.90; \\ Tiktok, score: 0.98\end{tabular} &
  Facebook, Tiktok: negative, score: 0.98 &
   &
   \\ \cline{1-3}
\begin{tabular}[c]{@{}l@{}}Apple's Stunning \$10 Billion \\ Blow to Facebook\end{tabular} &
  \begin{tabular}[c]{@{}l@{}}Apple, score: 0.99;\\ Facebook, score: 0.99\end{tabular} &
  Apple, Facebook: negative, score: 0.95 &
   &
   \\ \cline{1-3}
\begin{tabular}[c]{@{}l@{}}Why Apple's Privacy Changes \\ Hurt Snap and Facebook \\ but Benefited Google\end{tabular} &
  \begin{tabular}[c]{@{}l@{}}Apple, score: 0.99;\\ Snap, score: 0.99;\\ Facebook, score: 0.99;\\ Google, score: 0.99\end{tabular} &
  \begin{tabular}[c]{@{}l@{}}Apple, Snap: negative, score: 0.97;\\ Apple, Facebook: negative, score: 0.96;\\ Apple, Google: positive, score: 0.54\end{tabular} &
   &
   \\ \cline{1-3}
\begin{tabular}[c]{@{}l@{}}Apple and Google compete \\ against Facebook\end{tabular} &
  \begin{tabular}[c]{@{}l@{}}Apple, score: 0.99;\\ Google, score: 0.99;\\ Facebook, score: 0.99\end{tabular} &
  \begin{tabular}[c]{@{}l@{}}Apple, Google: neutral, score: 0.46\\ Apple, Facebook: negative, score: 0.70\\ Google, Facebook: negative, score: 0.64\end{tabular} &
   &
   \\ \cline{1-3}
\end{tabular}
\caption{Initial Results from the NER and ZSC models.}
\label{table_3}
\end{table}

\begin{table}[h]
\centering
\begin{tabular}{|l|l|p{7cm}|}
\hline
    \textbf{News headlines} &
    \textbf{GPT4 NER} &
    \textbf{GPT4 Explanations}
   \\ 
    \begin{tabular}[c]{@{}l@{}}Facebook Paid GOP Firm \\ to Malign Tiktok\end{tabular} &
    \begin{tabular}[c]{@{}l@{}}Facebook, \\ GOP Firm, and\\ Tiktok\end{tabular} &
    \begin{tabular}[c]{p{6.5cm}}
        Facebook and GOP Firm: The relationship appears to be positive from the context provided, as Facebook paid the GOP firm, indicating a transactional or business relationship.\\
        Facebook and TikTok: The relationship appears to be negative, as Facebook paid the GOP firm to ``malign" TikTok, indicating a harmful intent towards TikTok.\\
        GOP Firm and TikTok: The relationship can also be interpreted as negative, as the GOP firm is paid to ``malign" TikTok, suggesting they carried out actions intended to harm or discredit TikTok.
    \end{tabular}
   \\
    \begin{tabular}[c]{@{}l@{}}Apple's Stunning \$10 Billion \\ Blow to Facebook\end{tabular} &
    \begin{tabular}[c]{@{}l@{}}Apple, and\\ Facebook\end{tabular} &
    Apple and Facebook: The relationship appears to be negative, as the word ``blow'' suggests that Apple's action has led to a significant negative impact (a ``\$10 billion blow'') to Facebook.
   \\
    \begin{tabular}[c]{@{}l@{}}Why Apple's Privacy Changes \\ Hurt Snap and Facebook \\ but Benefited Google\end{tabular} &
    \begin{tabular}[c]{@{}l@{}}Apple,\\ Snap,\\ Facebook, and\\ Google.\end{tabular} &
    \begin{tabular}[c]{p{6.5cm}}
        Apple and Snap: The relationship appears to be negative, as Apple's privacy changes are stated to have ``hurt" Snap.\\
        Apple and Facebook: The relationship appears to be negative, as Apple's privacy changes are also stated to have ``hurt" Facebook.\\
        Apple and Google: The relationship appears to be positive, as Google is stated to have ``benefited" from Apple's privacy changes.
    \end{tabular} 
   \\
    \begin{tabular}[c]{@{}l@{}}Apple and Google compete \\ against Facebook\end{tabular} &
    \begin{tabular}[c]{@{}l@{}}Apple,\\ Google, and\\ Facebook\end{tabular} &
    \begin{tabular}[c]{p{6.5cm}}
        Apple and Facebook: The relationship appears to be negative, as Apple is mentioned as competing against Facebook.\\
        Google and Facebook: The relationship also appears to be negative, as Google is mentioned as competing against Facebook.\\
        Apple and Google: The relationship between Apple and Google is not directly mentioned in the headline, so it would be classified as unknown based on the information provided.
    \end{tabular}
    \\\hline
\end{tabular}
\caption{Explanations by GPT4.}
\label{table_4}
\end{table}

\section{Conclusion and Future Work}

In this work, we showed preliminary results on how foundation models (both instruction-tuned LLMs and NERs/ZSCs) can be used to create graphical abstractions that can be used for business decision making such as exploring new relationships, simulating network evolution, quantifying network effects and analyzing the impact of business decisions etc. We intend to improve our constructions by adding more context. We will compare our graph generation methodology with the current state-of-the-art approaches. Additionally, we aim to develop a model for predicting the relationships of unknown edges using social network tools such as Balance Structure Theory, as in Leskovec et al. (2010).

\printbibliography

\end{document}